\documentclass[12pt,a2paper]{article}
\usepackage{graphicx}
\usepackage{amsmath,amssymb}
\usepackage{color}
\usepackage[affil-it]{authblk} 
\usepackage{dsfont} 
\usepackage{subcaption}
\usepackage{amsthm}
\usepackage{array, tabu}

\def\R{\mathbb{R}}

\def\w{\mathbf{w}}
\def\x{\mathbf{x}}

\def\z{\mathbf{z}}
\def\1{\mathbf{1}}
\def\0{\mathbf{0}}
\def\P{\mathbf{P}}

\def\optlimits{\nolimits}

\newtheorem{theorem}{Theorem}

\begin{document}
\title{Discrete Potts Model for Generating Superpixels on Noisy Images}

\author[1]{Ruobing Shen%
  \thanks{Electronic address: \texttt{ruobing.shen@informatik.uni-heidelberg.de}}}
\affil[1]{Institute of Computer Science, Heidelberg University, Germany}

\author{Xiaoyu Chen}
\affil{Department of Computer Science, China University of Geosciences, China}
\author{Xiangrui Zheng}
\affil{Institute of Financial Studies, Shandong University, China}

\author[1]{Gerhard Reinelt%
  }

\maketitle

\begin{abstract}
Many computer vision applications, such as object recognition and segmentation, increasingly build on superpixels. However, there have been so far few superpixel algorithms that systematically deal with noisy images. We propose to first decompose the image into equal-sized rectangular patches, which also sets the maximum superpixel size. Within each patch, a Potts model for simultaneous segmentation and denoising is applied, that guarantees connected and non-overlapping superpixels and also produces a denoised image. The corresponding optimization problem is formulated as a mixed integer linear program (MILP), and solved by a commercial solver. Extensive experiments on the BSDS500 dataset images with noises are compared with other state-of-the-art  superpixel methods. Our method achieves the best result in terms of a combined score (OP) composed of the under-segmentation error, boundary recall and compactness.
\end{abstract}

\section{Introduction}
\label{sec:intro}
While the exact definition of a superpixel is vague, it is regarded as a group of perceptually meaningful connected regions of an image. A superpixel should contain pixels similar in color or other low-level properties, and therefore are likely to belong to the same physical world object. The concept of superpixels was introduced by \cite{super}, and was motivated by two aspects: firstly, pixels-grid is not a natural representation of visual scenes, but just a digital imaging ``artifact''; and secondly, the huge number of pixels in natural images prevents many computer vision algorithms being computationally efficient or even possible.

Computational efficiency comes from the reduction in the number of elements of a given image. Superpixels thus have been actively applied for a wide range of applications, and there exist many contributions, see \cite{SLIC,STUTZ2017} for an overview. Superpixels can be naturally obtained as the results of some image segmentation algorithms. To decrease the risk of the superpixels crossing object boundaries, these algorithms are applied in an over-segmentation mode. Examples are graph based\cite{Felzenszwalb2004}, and normalized cuts\cite{shi}.

Ideally, the following properties are desired for any superpixels algorithm:

\begin{itemize}
  \item Superpixels should adhere well to image boundaries.
  \item Superpixels should be regular in shape and size, with smooth boundaries.
  \item The number and size of superpixels should be controllable by the user.
  \item Superpixels should be non-overlapping and thus each pixel is assigned a label.
  \item Pixels with the same label (one superpixel) should be connected.
  \item Superpixels should have few parameters, so that it can be easily adjusted.
  \item Superpixels should be fast to generate.
\end{itemize}

Many superpixel algorithms produce regions of highly irregular shape and size. The boundaries are often highly irregular, since there is often no explicit constraint or penalty on the boundary length. There are advantages of superpixels with regular shapes and sizes. Apart from visual appealing, in case a superpixel does cross the boundary, since the size is controlled, the error rate as well. Also, given the
same amount of superpixels, more regular and compact size
would lead to fewer edges and thus a sparser graph,

\begin{figure}[t]
  \center
 \includegraphics[width=0.8\columnwidth]{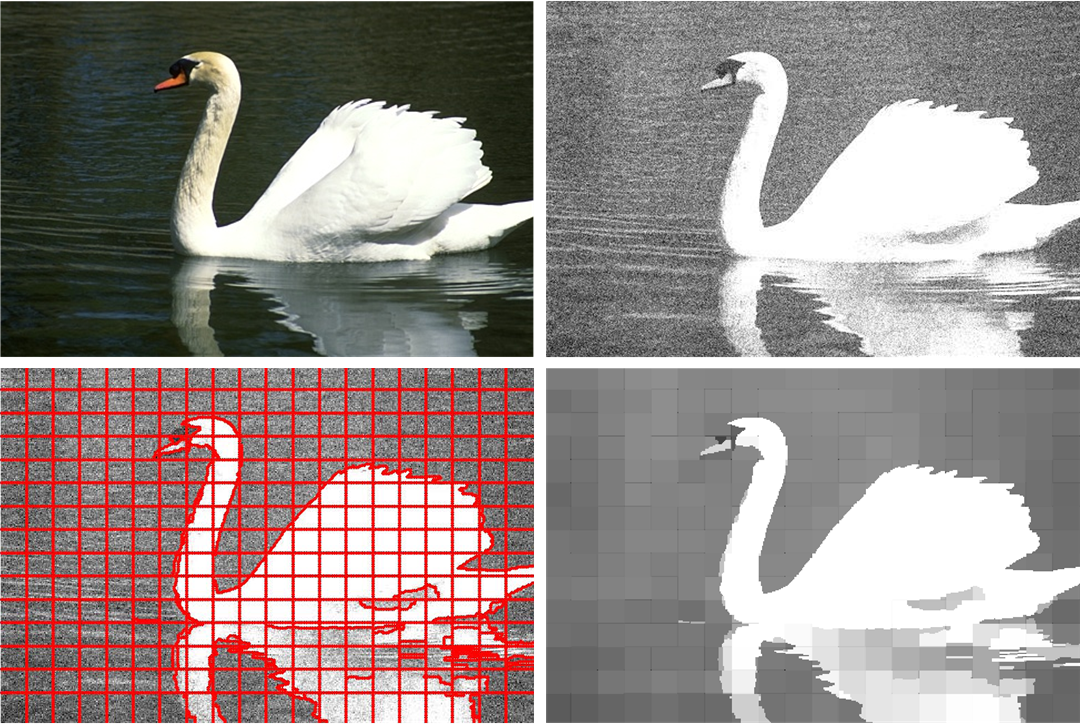}
 \caption{Top left: RGB image. Top right: gray-scale image with Gaussian noise. Bottom left: superpixel results of PMcut. Bottom right: denoised output.}
 \label{fig:cover}
\end{figure}

\textbf{Related Works.} We review some state-of-the-art superpixel algorithms, which are either highly ranked in a recent survey paper \cite{STUTZ2017} or give regular shapes.

\textbf{CW --} Compact Watershed \cite{cw} is based on SLIC and~\cite{water}, and contains two modified algorithms. The first one speeds up SLIC, while the second creates uniformly shaped superpixels as opposed to \cite{water}. It is computational very efficient, takes only $10$-$33$ ms for segmentation an image of size $ 321\times 481$.

\textbf{SEEDS --} Superpixels Extracted via Energy-Driven Sampling~\cite{seeds} proposed an approach based
on a simple hill-climbing optimization. It starts from an initial superpixel
partitioning (e.g., rectangular patches), then continuously refines the superpixels by modifying
the boundaries. The energy function is based on enforcing color similarity between the boundaries and the
superpixel color histogram. It can control the number of superpixels but not the compactness. 

\textbf{SLIC --} Simple Linear Iterative Clustering~\cite{SLIC} uses k-means clustering algorithm. It initializes by seeding pixels as cluster centers, and uses color and spatial information for updating. Post-processing is needed to ensure connectivity of superpixels. This approach offers control over the number of superpixels and the compactness. 

\textbf{ETPS --} Extended Topology Preserving Segmentation~\cite{etps} partition the image into a regular grid as initial superpixel segmentation, and then exchanges pixels between neighboring superpixels iteratively. It uses a coarse-to-fine energy update strategy, and uses block coordinate gradient descent to minimize the energy. However, it produces superpixels with irregular sizes and shapes.

While previous superpixel algorithms may work well on clean images, they do not have a systematical way of dealing with noisy images, thus may suffer from the presence of noise. Also, there has not been a comprehensive study on superpixels algorithms applied to noisy images.

Our method combine segmentation and denoising in one framework. The generated superpixels are connected and the image is denoised as a by-product (see Fig.~\ref{fig:cover}). 
It is worthwhile to highlight the contribution of this paper:

\begin{itemize}
  \item We conduct comprehensive experiments against $4$ superpixels algorithms on the BSDS500 dataset images with noises.
  \item Our method (\textbf{PMcut}) produces connected and  compact superpixels.
  \item PMcut has only three parameters (one sets the number of superpixels and meanwhile controls the maximum superpixel size, one penalizes the boundary length, one sets the time limit).
  \item PMcut achieves the state-of-the-art on noisy images, even against other algorithms on denoised images.
\end{itemize}

\section{Potts Model with Multicut Constraints}
\label{sec:1D}
In this section, we give a detailed description of our superpixel method (PMcut). We first review the discrete Potts model and the multicut problem, then we introduce our MILP formulation for simultaneous image segmentation and denoising. We conduct some preliminary experiments on PMcut with toy examples (of size $40\times 40$). Finally, we discuss how PMcut can be applied to generate superpixels of the input image.

We assume the input image has gray-scale values (RGB images can be easily transformed) for pixels located on an~$m\times n$ grid.
For representing relations between neighboring pixels, we define the corresponding grid graph $G=(V,E)$ where $V=\{(1,1),\ldots, (m,n)\}$ denote the set of pixels, and $E$ contains edges between pixels which are horizontally and vertically adjacent. A segmentation of the image in our setting is a partition of $V$ into sets $\{V_1, V_2, \ldots, V_k\}$ such that $\cup_{i=1}^kV_i=V$, and $V_i \cap V_j=\emptyset$, $\forall i\neq j$. In other words, the image segmentation problem corresponds to a graph partitioning problem.

\begin{figure}[t]
 \centering
 \includegraphics[height=4.5cm]{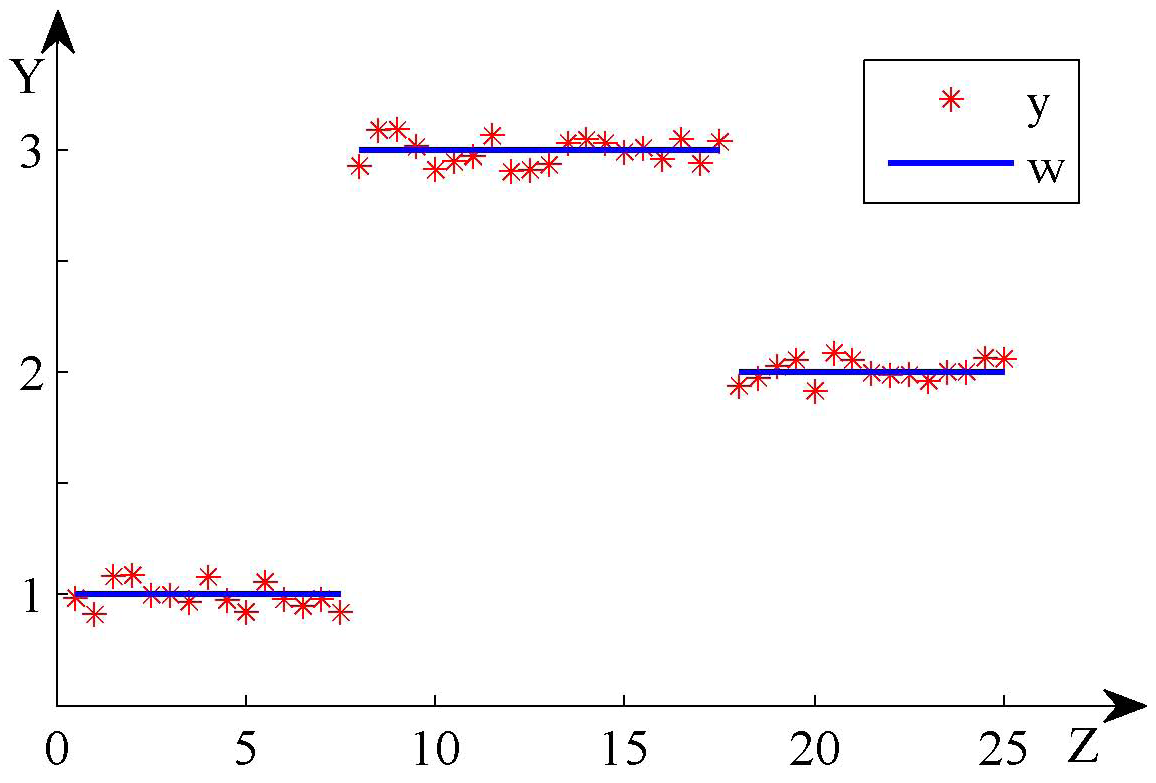}
 \includegraphics[height=4.5cm]{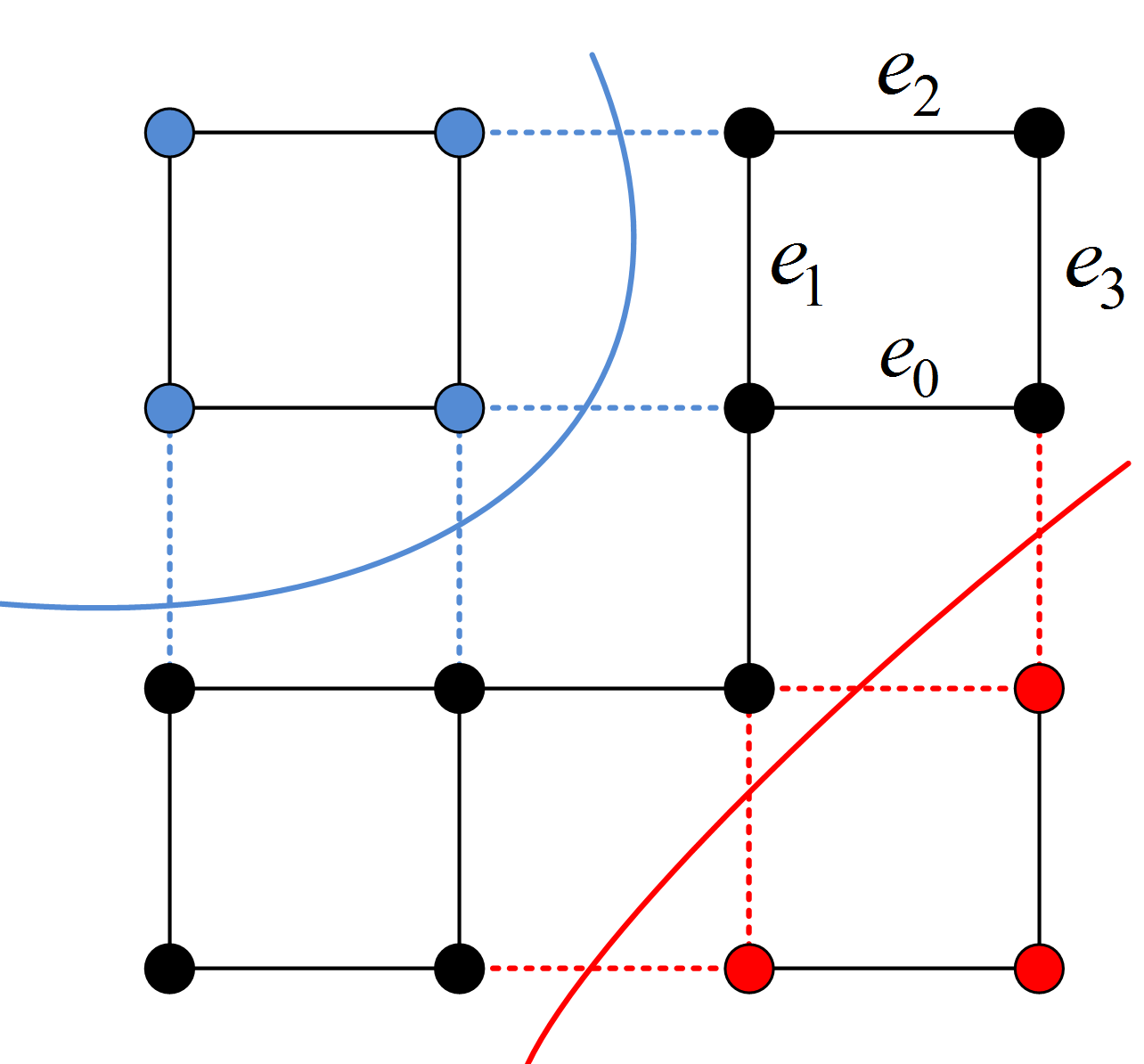}
 \caption{Left: 1D Potts model, 3 segments and 2 active edges. Right: multicut in a $4\times4$-grid image with 3 segments.}
 \label{fig:multicut2}
\end{figure}

\subsection{Discrete Potts Model}
The main tools for denoising are variational methods like
the approach with Potts priors~\cite{potts}, which was designed to preserve sharp discontinuities (edges) in images while removing noises. While in this paper, we point out that the discrete Potts model can be also applied for image segmentation.

Denote the input intensity values (color or gray-scale value) of pixels as $Y=(y_{ij})$. Since we are working on a grid graph, we denote the $i$-th row of $Y$ by~$Y^r_i$ and the $j$-th column by~$Y^c_j$. Let $W=(w_{ij})\in \R^{m\times n}$ be the fitting (denoised) value of pixels (Similarly we have $W^r_i$ and $W^c_j$). We divide $E=E^r\cup E^c$ into its horizontal (row) edges $E^r$ and its vertical (column) edges~$E^c$. The classical (discrete) Potts model has the form
\begin{alignat}{2}
\min_{W} \;\sum\optlimits_i\|W_i^r-Y_i^r\|_k + \sum\optlimits_j\|W^c_j-Y^c_j\|_k +\lambda\|\nabla^1 W^r\|_0+\lambda\|\nabla^1 W^c\|_0, \label{TV}
\end{alignat}
where the first two terms measure the $\ell_k$ norm difference between every row and column vector of $W$ and~$Y$, and the further terms measure the number of oscillations in $W$ per row and column. Recall that the \emph{discrete first derivative} $\nabla^1 \w$ of a vector~$\w\in\R^n$ is defined as the $n-1$ dimensional vector $(w_2-w_1,w_3-w_2,\ldots,w_n-w_{n-1})$ and the $\ell_0$ norm of a vector gives the number of its nonzero entries. The scalar~$\lambda$ is a parameter for regularization.

In general, solving the discrete Potts model \eqref{TV} is $\mathcal{NP}$-hard, and~\cite{Fast} uses local greedy methods. Recently, a MILP formulation is applied in~\cite{bertsimas2016} to deal with the~$\ell_0$ norm for a similar problem in statistics called the \emph{best subset selection problem}.

\subsection{MILP formulation of the Discrete Potts}
Since $G$ in $2$D is a grid graph, $E^r=\{E^r_1, E^r_2, \ldots, E^r_m\}$, where $E^r_i$ is the set of horizontal edges in the $i$-th row. Similarly, we have $E^c=\{E^c_1, E^c_2, \ldots, E^c_n\}$. We propose to formulate~problem \eqref{TV} as a MILP by introducing $m\cdot (n-1)+(m-1)\cdot n$ binary edge variables~$x_e$. 
We would like the following properties to hold
 
\begin{align}
	\nabla^1 w_e= 0  &\Leftrightarrow x_{e} = 0, \; \; \forall e\in E,\nonumber\\
	\nabla^1 w_e\ne 0  &\Leftrightarrow x_{e} = 1, \; \; \forall e\in E,\label{logic}
\end{align} 
where $\nabla^1w_e:= w_{h_e}-w_{t_e}$, and $h_e$, $t_e$ denotes the two end nodes of any edge $e\in E$. 
We call an edge $e\in E$ \emph{active} if $x_e=1$, otherwise it is \emph{dormant}.
 
In $1$D, if the above condition holds, nodes within two active edges will have the same intensity, while  the two end nodes of an active edge have different. It thus corresponds to a piecewise linear constant fitting problem, see left part of Fig.~\ref{fig:multicut2} for an example. 


A mixed integer programming (MIP) formulation for \eqref{TV} that satisfies properties~\eqref{logic} reads
\begin{alignat}{2}
	\min \;\; \sum\optlimits_{i=1}^m\sum\optlimits_{j=1}^n&|w_{ij}-y_{ij} |+\lambda \sum\optlimits_{e\in E} x_{e}\label{eq:Pott_1D2}\\ 
	|\nabla^1 w_e| &\leq M x_{e}, \quad\forall e\in E^r_i, \tag{\ref{eq:Pott_1D2}a}\label{bigmcons}\\
	|\nabla^1 w_e| &\leq M x_{e}, \quad\forall e\in E^c_j, \tag{\ref{eq:Pott_1D2}b}\label{bigmcons2}\\
	x_e &\in \{0,1\},\;\;e\in E, \tag{\ref{eq:Pott_1D2}c}\\
	W &\in \R^{m\times n}\nonumber\tag{\ref{eq:Pott_1D2}d},
\end{alignat}
where $[m]$ denotes the discrete set $\{1,\ldots,m\}$ and $M$ should be big enough so that~(\ref{bigmcons},\ref{bigmcons2}) are always valid when $|\nabla^1 w_e|\neq 0$. Note that we use the $\ell_1$ norm over $\ell_2$ because it is more robust to outliers. Secondly, it is easier to 
model the $\ell_1$ with linear constraints, where constraint~\eqref{bigmcons} is replaced by the two constraints
$\nabla^1 w_e \leq Mx_{e}$ and $-\nabla^1 w_e \leq Mx_{e}$. The term
$|w_{ij}-y_{ij}|$ is replaced by $\varepsilon_{ij}^+ +\varepsilon_{ij}^-$ where $w_{ij} - y_{ij} = \varepsilon^+_{ij} - \varepsilon^-_{ij}$
and $\varepsilon_{ij}^+, \varepsilon_{ij}^- \geq 0.$  In this paper, for simplicity,
we will just specify models in form~\eqref{eq:Pott_1D2}.

\begin{theorem}
The optimal solutions of~\eqref{eq:Pott_1D2} satisfies properties~\eqref{logic}.
\end{theorem}
The proof can be found in the supplementary materials.
\subsection{The Multicut Problem}
\label{sec:multicut}
The multicut problem (Multicuts) in~\cite{prob} formulates the image segmentation problem as an edge labeling problem. For a partition ${\cal V}=\{V_1, V_2, \ldots, V_k\}$ of $V$, the edge set $\psi(V_1, V_2, \ldots, V_k) = \{uv \in E \mid \exists i \ne j \text{ with } u\in V_i \text{ and } v \in V_j\}$ is called the \emph{multicut} induced by $\cal V$. Similar to the previous section, we introduce binary edge variables $x_e$  and the multicut is just a set of active edges.

With edge weights $c: E\rightarrow \R$ typically representing differences between pixels' intensities, the multicut problem for the unsupervised image segmentation~\cite{kappesg} can be formulated as the following integer linear program~(ILP)
\begin{align}
\min \;\; \sum\optlimits_{e\in E}& -c_ex_e + \sum\optlimits_{e\in E}\lambda x_e\label{multicut}\\
\sum\optlimits_{e\in C\setminus\{e'\}}x_e&\geq x_{e'}, \;\;  \forall \;\text{cycle} \;C\subseteq E, e'\in C,\tag{\ref{multicut}a} \label{multicut_a}\\
x_e &\in \{0,1\},\;\; \forall e\in E, \tag{\ref{multicut}b}
\end{align}
where $\lambda$ serves as the penalty term for the number of active edges. Constraints~\eqref{multicut_a} are called the \emph{multicut constraints} and they enforce the consecutiveness of the active edges, and thus the connectedness of each segment. 
Thus, a segment is composed with the maximal set of vertices induced only by the dormant edges. See the right part of Fig.~\ref{fig:multicut2}, where the dashed edges form the multicut.

It is well known that if a cycle $C\in G$ is chordless,
then the corresponding constraint~\eqref{multicut_a} is facet-defining for the multicut 
polytope~\cite{kappesg}. However, the number of such inequalities is exponentially in $|V|$, and they are usually added iteratively using a cutting plane method.

\subsection{Adding additional facet-defining constraints}
\label{redun}
It is common practice to add additional constraints to a MIP for computational efficiency. A \emph{violating constraint} ``cuts off'' feasible solutions of a formulation, which are not optimal. A
constraint is \emph{redundant} if it is not necessarily needed for a formulation to be valid. However, they may be useful because they forbid some fractional solutions during the branch-and-bound approach, where the MIP solver iteratively solves the linear programming (LP) relaxation.

Let $S:=\{(\w,\x)\;\vert\; (\w,\x) \text{ is feasible to  problem~\eqref{eq:Pott_1D2}}\}$ be the set of all feasible solutions to~\eqref{eq:Pott_1D2}, and $S_I:=\{\x\subset (\w,\x)\;\vert\; (\w,\x)\in S\}$ denotes its projection to~$\x$. Denote $S^o\subseteq S$, where $(\w, \x)\in S^o$ satisfies property~\ref{logic}, and we let $S_I^o$ be the projection of $S^o$ to $\x$.

\begin{theorem}
The multicut constraints~\eqref{multicut_a} are redundant for $S_I^o$ of~\eqref{eq:Pott_1D2}.
\end{theorem}
The proof can be found in the supplementary materials.

Note that the multicut constraints~\eqref{multicut_a} are not always valid for the feasible set of~\eqref{eq:Pott_1D2}, one example is shown in the supplementary materials.
Since we know the multicut uniquely defines an image segmentation, we can now conclude that the optimal solution of the discrete Potts model~\eqref{eq:Pott_1D2} also defines a unique segmentation.

\begin{theorem}
The chordless multicut constraints are facet-defining for the convex hull of $S_I^o$.
\end{theorem}
The proof can again be found in the supplementary materials.

Inspired by the multicut problem~\eqref{multicut}, in a grid graph, although the number of multicut constraints is still exponential, it may be advantageous to add the following $4$-edge chordless cycle (e.g., cycle $e_0$-$e_1$-$e_2$-$e_3$ in Fig.~\ref{fig:multicut2}) constraints

\begin{equation}
\sum\optlimits_{e\in C\setminus\{e'\}}x_e\geq x_{e'}, \;\;  \forall \;\text{cycles $C\subseteq E$, $|C|=4$, $e'\in C$} \label{multicut4}
\end{equation}
to \eqref{eq:Pott_1D2}.
Because it is the simplest (in terms of the number of edges) chordless cycle in a grid graph, plus the number of such constraints is only linear in $|E|$.

Of course, there exists more complicated strategy on selecting such cuts, which is beyond the focus of this paper. Interested readers may refer to~\cite{Sontag2012,Johnson2008,Batra2011} for more information. 
We will show experiments in Sec.~\ref{sec:experiment} on the effects of adding the above constraints.

\subsection{Main formulation: adding multicut constraints to Potts}
\label{sec:2dproblem}
The main formulation (PMcut) is obtained by adding \eqref{multicut4} into \eqref{eq:Pott_1D2}, which could be seen as an extension of~\cite{BU17} from discrete to continuous labeling space (if we treat $w_i$ as the label of node $i$).
Note that by placing penalty term $\lambda$ on each actives edge, it in turn penalizes the length of segment boundary, thus encouraging compact shape of the segmentation. Upon solving \eqref{eq:Pott_1D2}, variable $x_e$ gives the boundaries of each segment, and $w_{ij}$ is the denoised value for every pixel. Thus, we get both segmentation and denoising of the input image.

PMcut is then solved using any standard off-the-shelf MIP solver, where advanced branch-and-cut methods are employed. One can set limits on the running time or the MIP gap. The MIP solver could return the best feasible solution found and the MIP gap upon termination. 
\subsection{Preliminary Experiments}
\label{sec:experiment}
All the computational experiments in this paper were performed using a commercial MIP solver Gurobi $7.5.1$, on an Intel i5-4570 quad-core desktop, with 16GB RAM. If the input image is RGB, we first transform it into gray-scale valued image, and normalize the intensity of each pixel to $[0,1]$.

Note that we choose to work on gray-scale mainly due to computational reasons. In theory, PMcut can be easily adapted to RGB images, by introducing continuous fitting variables $w$ for every channel. But this will trigger $O(mn)$ more variables and constraints.

For visualization, we take an image from~\cite{martin}, resize it to $40\times 40$, and add Gaussian and Salt and Pepper (S\&P) noise with a time limit of $50$ secs.

\textbf{Parameter setting.}
We simply set the constant $M$ to be $1$, so that~(\ref{bigmcons},\ref{bigmcons2}) are always valid. We compute the average intensity of each $5\times 5$ pixel block of the image, and then calculate the absolute difference of their maximum and minimum value, denoted~$Y^{*}$. So~$Y^{*}$ somehow represents the global contrast of the image. Since we assume the input image contains noise, the regularization parameter is set to $\lambda=\frac{1}{4}\sigma Y^*$, where $\sigma$ is a user-defined parameter. When there exists an outlier, PMcuts will not treat the outlier as a one-pixel segment, since doing so would incur the penalty $4\lambda$ (equals $\sigma Y^*$).

\begin{figure}[t]
  \center
 \includegraphics[width=0.9\columnwidth]{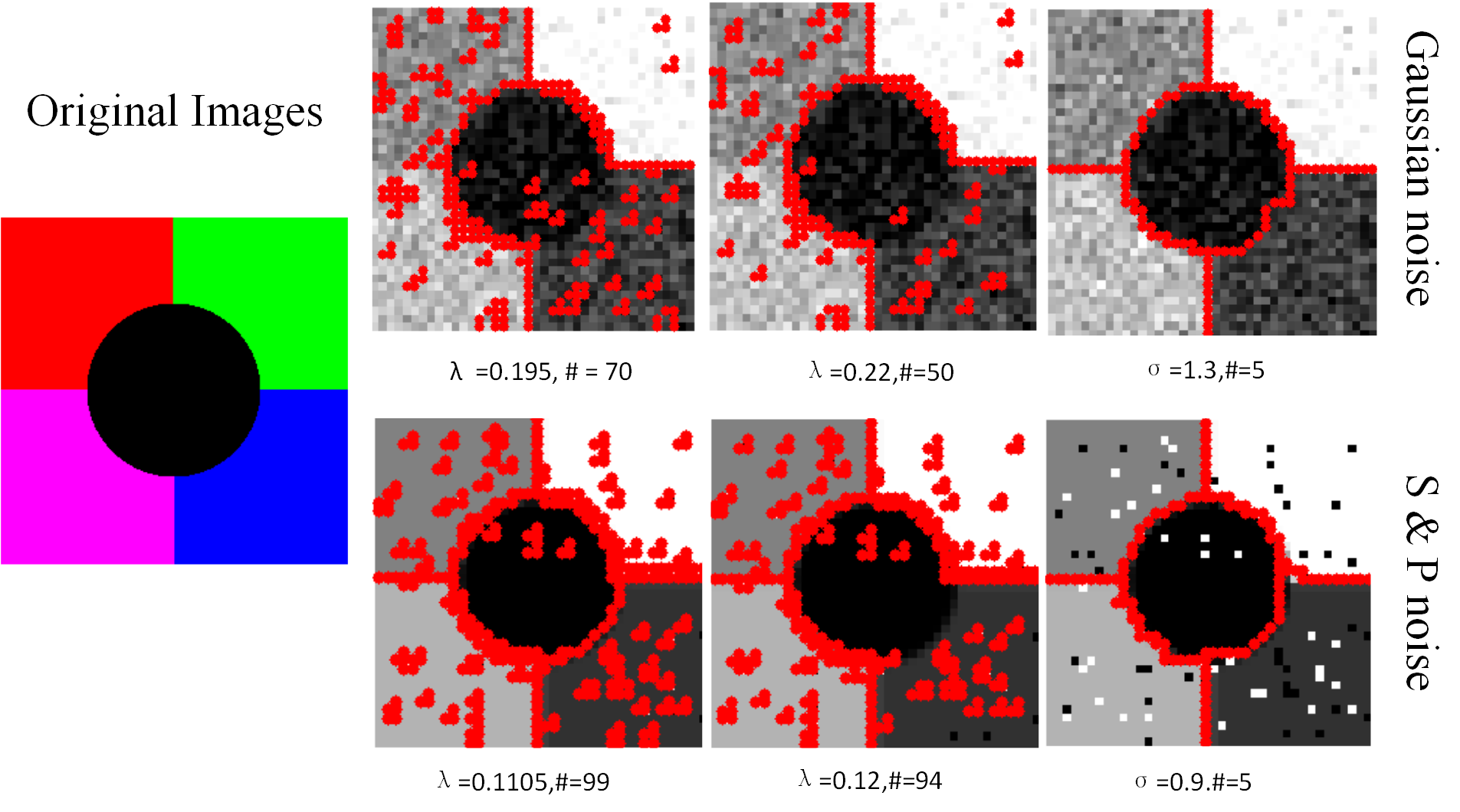}
 \caption{Segmentation results using Multicuts and PMcuts. $\lambda$ and $\sigma$: parameters.}
 \label{fig:ex}
\end{figure}

\textbf{Multicuts VS PMcut.}
The right part of Fig.~\ref{fig:ex} shows the input image, the detailed setting of the parameters and segmentation results. The first two columns are Multicuts with different $\lambda$. The third column represents PMcut. The number of segments is reported in Fig.~\ref{fig:ex}. Multicuts takes less than one sec in all three instances. Although found optimal solution of the first instance in less than 6 sec, PMcut did not converge in the second instance with the time limit. The MIP gap is $1.91\%$.

\textbf{With and without constraints~\eqref{multicut4}.} We report that with~\eqref{multicut4}, PMcut saves $0.9$ secs in the first instance, and narrows down $23.5\%$ of the MIP gap in the second one, compared to without~\eqref{multicut4}.
We conduct $10$ more experiments (size $40\times 50$, with $5\%$ $S\&P$ noise, time limit $50$ secs), and report that these constraints help reduce the MIP gap from $66\%$ to $37\%$ on average.

We can conclude that it is hard to control the desired number of segments for Multicuts.
On the other hand, although requiring more computational time, PMcut is robust to noise, easier to control parameters, and give better segmentation results. Also, it is clearly beneficial to add constraints~\eqref{multicut4} to the Potts~\eqref{eq:Pott_1D2}, as we did in PMcut.

\subsection{Our proposed superpixel algorithm}
Our superpixel algorithm is quite straightforward. We begin by first decomposing the input image into $K$ rectangular patches, where $K$ is a user defined parameter approximating the number of desired superpixels. We then apply PMcut within each image patch. The superpixel is also upper bounded by the size of the patch. 

Since PMcut is $\mathcal{NP}$-hard, smaller patches are desired for the sake of computational efficiency. In practice, we notice that with the increase of the number of patches, the overall computational time decreases since the patch size shrinks.

Note also that every image patch can be computed separately, thus parallelism can be fully adopted. In practice, one can also set a MIP gap threshold for PMcut within each patch. As a matter of fact, we often notice the segmentation results are already quite good even though it has not reached the optimal solution. Hence it makes sense to create a stopping criterion when PMcut either hits a time limit or it finds a solution within a certain MIP gap. We set the threshold of MIP gap to $2\%$ in later experiments.

\section{Computational Experiments}
\label{sec:experiment}
In this section, we present a detailed evaluation of our superpixel algorithm against the other state-of-the-art algorithms on noisy images, with respect to superpixel sizes of $600$, $1200$, $1800$ and $2400$.  The BSDS500 consists of $500$ images (size $321\times 481$) splitting into $200$ training, $100$ validation and $200$ test images. We conduct extensive parameter training before testing. For testing, we choose $100$ test images to add $0.3$ Gaussian noise, and the rest $100$ to add $15\%$ $S\&P$ noise. The test sets are thus divided according to these two types of noises.

\subsection{Evaluation Benchmark}
Among other metrics, the Under-segmentation Error~(\textbf{UE}), Boundary Recall (\textbf{Rec}) and Compactness~(\textbf{CO})~\cite{schick2012,STUTZ2017} are probably the most widely used ones. The former two are standard measures for segmentation boundary adherence.

Given a superpixel segmentation and a ground truth, UE measures the fraction of superpixels that ``leak'' across the boundary of a ground truth segment. We adopt the updated formulation of UE in \cite{STUTZ2017}, that does not over-penalize superpixels which only overlap slightly with ground truth segment. Overall, under any formulation of UE, the lower score is better.

Rec measures what fraction of the ground truth edges fall within a local neighborhood of size $k\cdot k$ of a superpixel boundary. Like in~\cite{STUTZ2017}, we set $k=3$. As superpixels are expected to adhere to boundaries, high Rec score is desirable.  

CO was introduced to evaluate the compactness of superpixels. It compares the area $A(S_i)$ of each superpixel $S_i$ with the area of a circle with the same perimeter $P(S_i)$. The higher CO is better.

The detailed formulations of the UE, Rec and CO scores can be found in the supplementary materials. It is shown in~\cite{STUTZ2017} that there is a trade-off between Rec and CO, and higher CO usually results in lower Rec. 
 
\subsection{Parameter optimization and post-processing}
For the sake of fair comparison, we optimized parameters for the competing algorithms on $100$ training images where half are added with Gaussian noise and another half with $S\&P$ noise. The noise level is the same with testing images.
We conduct discrete grid search, jointly optimizing UE, Rec and CO, i.e., OP=$0.4*$UE+$0.4*$Rec+$0.2*$CO. 
For implementation details, please refer to~\cite{STUTZ2017}.

Since some superpixel algorithms do not ensure connectivity, a connected component algorithm is designed by~\cite{STUTZ2017}, which results in many tiny superpixels. Hence additional merging algorithm is designed to merge tiny superpixels into larger neighboring ones. We set the minimum threshold (number of pixels in any superpixel) to be $10$. 

We adopt the code from~\cite{STUTZ2017} for both parameter optimization and post-processing, for implementation details, please refer to~\cite{STUTZ2017}. While PMcut did not train the parameter $\sigma$ (since it is computationally very expensive, and we set it to $0.5$), and it works on the gray-scale level,  we show that it already achieves the best $OP$ score on all experiments.

\subsection{Quantitative Comparison}
Rec and UE offer a ground truth dependent overview to evaluate the performance of superpixel algorithms, while CO is independent of ground truth. For the former two, the superpixel results are compared with the provided five ground truth segmentations per image, and we choose the best and average out of $5$. We then compute the average score of all test images.

\begin{figure}[t!]
  \centering
    \begin{subfigure}{0.95\columnwidth}
                    \includegraphics[width=.99\columnwidth]{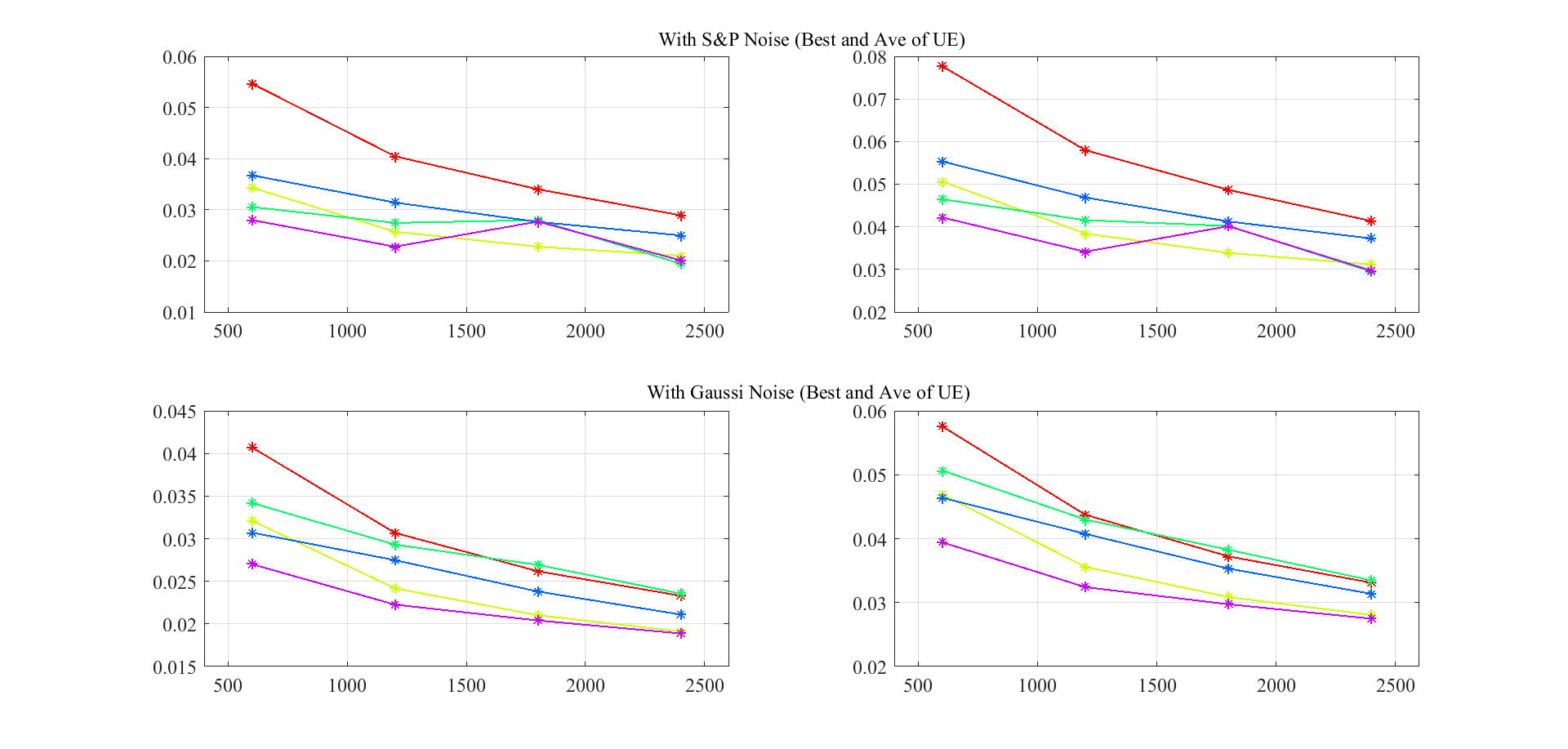}
    \end{subfigure}\\
    \begin{subfigure}{0.95\columnwidth}
            \includegraphics[width=.99\columnwidth]{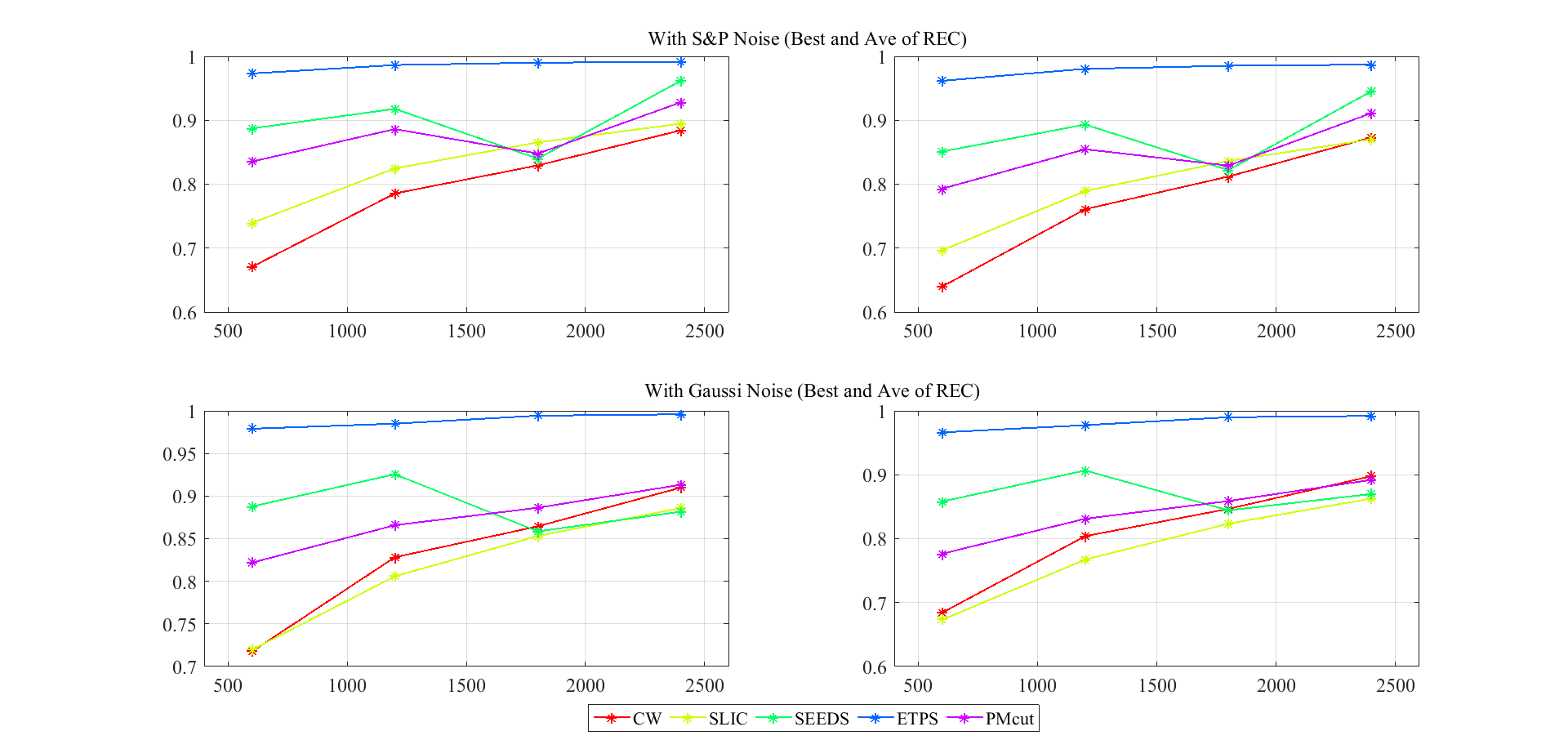}     
    \end{subfigure}       
 \caption{Top $2$ rows: UE score (lower is better), bottom $2$: Rec score (higher is better).
 X-axis: number of superpixels, Y-axis: score. }
 \label{fig:rec}
\end{figure}

\textbf{UE}. Top $2$ rows of Fig.~\ref{fig:rec} plots the UE score of each method against the increasing number of superpixels. The first row depicts results on the $100$ images with $S\&P$ noise, and the second row another $100$ images with Gaussian noise. PMcut is the clear winner on UE score with SLIC comes next place, while CW is the clear loser. There is a clear tread that with increasing number of superpixels, UE score gets better.


\textbf{Rec}. Rec score of each method is plotted similarly in the bottom $2$ rows of Fig.~\ref{fig:rec}.
On both kinds of noisy images, we can easily identify ETPS is the top performing method.
SEEDS and PMcut come the next places, while there is no clear loser. There is also a tread that with increasing number of superpixels, REC score gets better, with the exception on SEEDS.

\begin{figure}[t]
  \center
 \includegraphics[width=0.95\columnwidth]{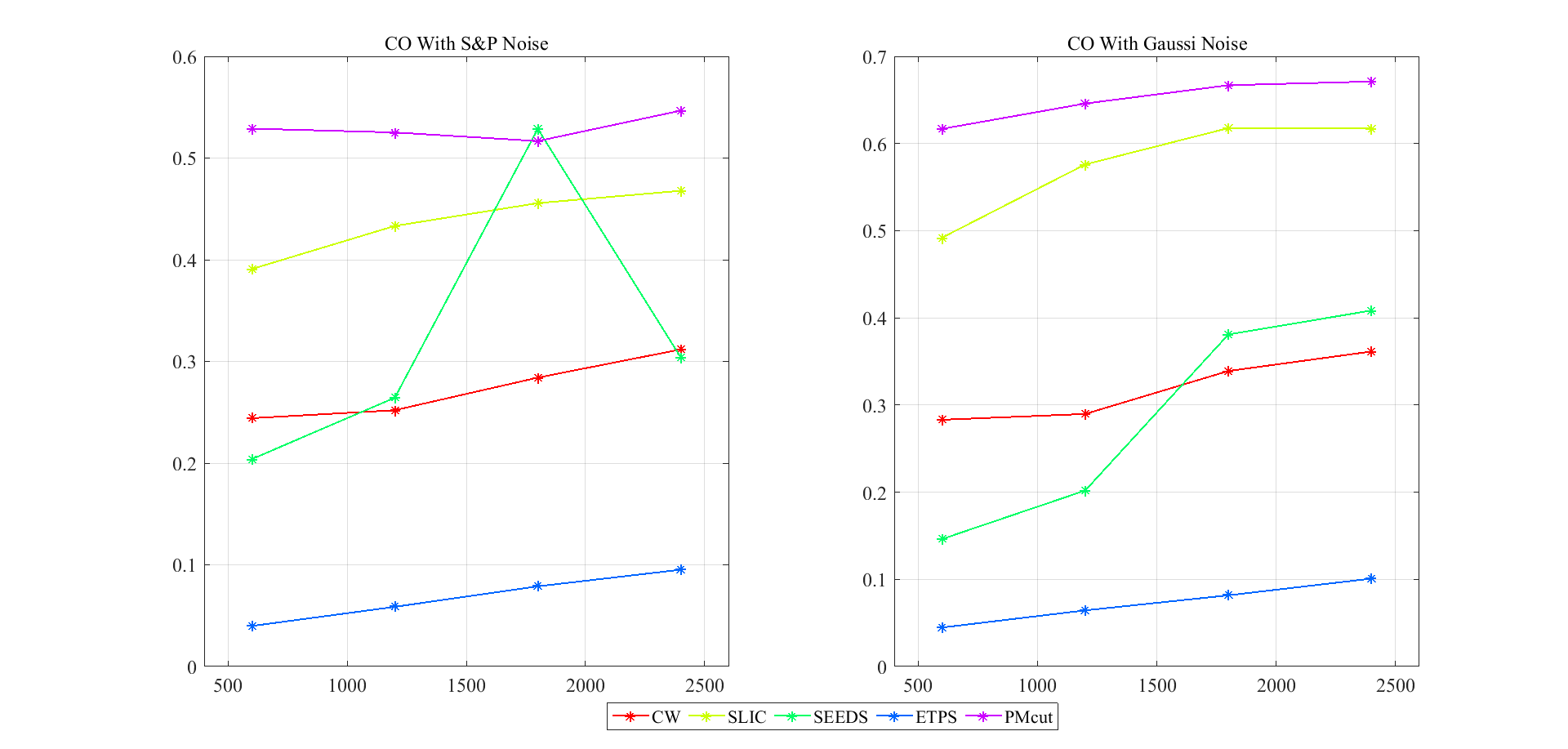}
 \caption{CO score of each method, higher is better.}
 \label{fig:co}
\end{figure}

\textbf{CO}. Fig.~\ref{fig:co} shows the CO score. In comparison with other algorithms, PMcut shows a significant advantage, and ETPS has the worst score. While ETPS has the best Rec score, this again demonstrates the trade-off between Rec and CO.

\textbf{OP}. Since all the competing algorithms are trained towards OP, it is thus the most important score in this paper. We report that PMcut achieves the $1st$ place on all $24$ scenarios. We depict one table of OP scores that contains $1$ (out of $2$) noise type, $1$ (out of $4$) superpixel size and $3$ ways of computing score in the supplementary materials. Table~\ref{table:op} shows the average OP score amongst all.

\begin{table}[h]
\centering
\begin{tabu} { | X[c] | X[c] | X[c]   | X[c] | X[c] |}
 \hline
 CW & SLIC & SEEDS  & ETPS & PMcut \\
 \hline
 0.75  & 0.80 & 0.79 & 0.79 & 0.84 \\
\hline
\end{tabu}
\caption{Average OP score of $5$ superpixel methods.}
\label{table:op}
\end{table}

\textbf{Time}. While all the other superpixel algorithm takes less than $1$ second, the run time of PMcuts is doomed by the $\mathcal{NP}$-hard nature of the MILP. Among all, CW and PB are the fastest, on average take only $0.01$ sec.

We set a time limit of each rectangular patch of PMcut to be $2$, $0.5$, $0.1$ and $0.05$ sec, for the desired superpixel size (also the number of MILPs) of $600$, $1200$, $1800$ and $2400$ respectively. The corresponding total average running time for the whole image is $272$, $134$, $40$ and $27$ secs. Finally, the average Gurobi MIP gap is $12.2\%$, $9.8\%$, $10.3\%$ and $9.8\%$, respectively.
 
One can imagine that as the number of rectangular patches increases, the total time as well as the average MIP gap would decrease. In addition, parallelism can be fully adopted to speed up computation.

\subsection{Qualitative Comparison}
\begin{figure}[t!]
\begin{center}
\includegraphics[width=.99\linewidth]{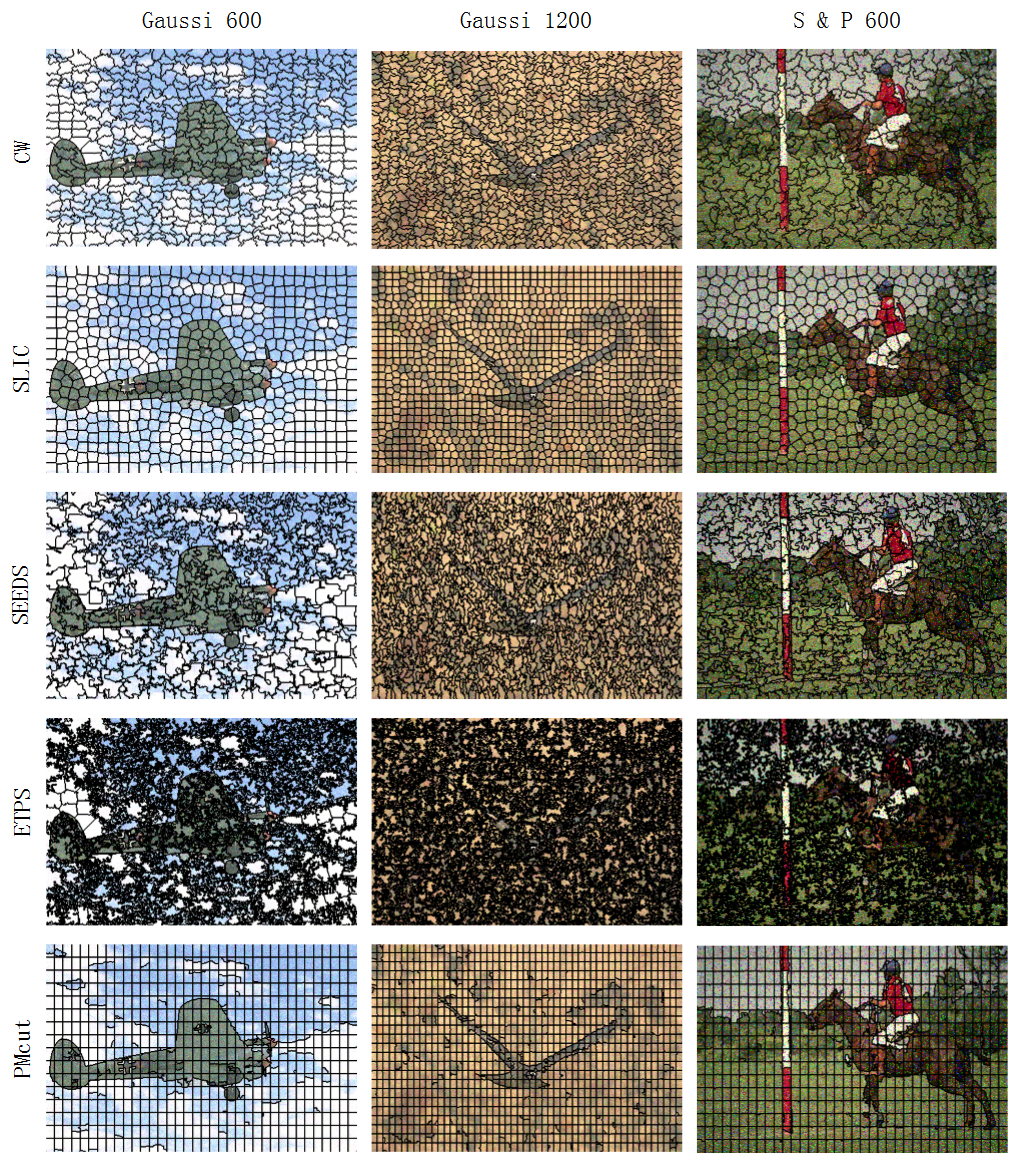}
\end{center}
   \caption{Superpixels using $5$ methods on images with Gaussian and $S\&P$ noise.}
\label{fig:compare}
\end{figure}

Visual quality can be determined by considering compactness, regularity and smoothness. Here, regularity corresponds to both the size and the arrangement of each superpixel, and smoothness refers to the superpixel’s boundary.
Fig.~\ref{fig:compare} shows superpixel results on $3$ images (with two kinds of noises) using all methods, with approximate superpixel number equals $600$ and $1200$. Note that these images are intended to be representative, but superpixel segmentations may vary across different images.

Most of the algorithms provide solid adherence to salient image boundaries, especially when the number of superpixels is large. Compactness and smoothness vary a lot across algorithms and a compactness parameter is beneficial to control the degree of compactness which allows to trade boundary adherence for CO. Ideally, compactness should be increased while only slightly sacrificing Rec.

At a first glance, the superpixels generated by PMcut, SLIC are visually more appealing.
Superpixels of PMcut are initialized and bounded by each rectangular patch. Since most of them consist of only background, the resulting superpixel fill in the whole rectangular patch. Hence most of the superpixels are compact and regular in size. PMcut also benefits from the penalty term $\lambda$ on the boundary length, thus creating smooth superpixels. It happens that SLIC loses track of some boundary details, e.g., the helmet in the third image. 

In contrast, although being able to capture most of the important boundaries (with high Rec), ETPS and SEEDS produce highly irregular and non-smooth superpixels. Their superpixel size varies a lot, with many tiny superpixels and yet huge ones. One can hardly recognize the original image, if the number of superpixel is large, e.g., ETPS in the second image.

In conclusion, we find that the evaluated ETPS and SEEDS show inferior visual quality. On the other hand, PMcut and SLIC demonstrate good visual quality at the expense of missing several boundaries. Overall, PMcut and SLIC achieve good balance between accuracy and visual appearance.

\subsection{Competing algorithms applied on denoised images}
\begin{figure}[t]
  \center
 \includegraphics[width=0.85\columnwidth]{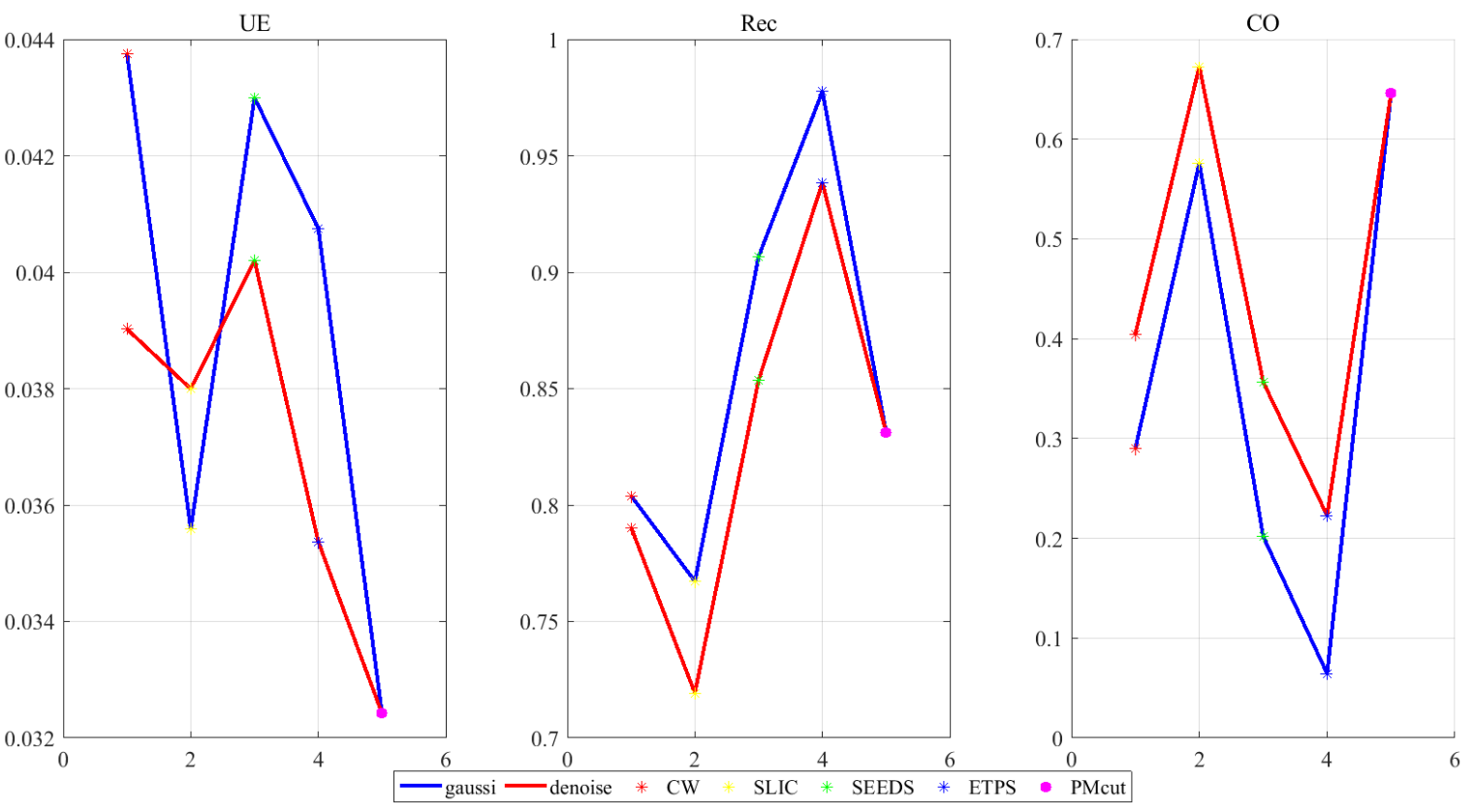}
 \caption{Comparing superpixel algorithms applied on Gaussian and denoised images (PMcut only on noisy images), number of desired superpixels: 1200}
 \label{fig:gaussian}
\end{figure}

We conduct additional experiments by first applying~\cite{non-local2005} (we set the two parameters $\sigma_s = 5$ and $h = 0.55\sigma_n$, where $\sigma_n$ is the noise variance of $n$ pixels.) to denoise the $100$ images with Gaussian noise. The competing superpixel algorithms are then run on the $100$ denoised images, then being compared to PMcut directly applied to the noisy images. The result is shown in Fig.~\ref{fig:gaussian} against on the noisy images, where $\#=1200$ and only the best of $5$ ground-truth score is depicted. As one can notice, all algorithms actually performs worse on the Rec score, while better on the CO score. We think it may be due to the fact that the denoising algorithm smooths the boundary too much. The average running time for the sophisticated denoising algorithm is over $60$ secs per image, hence our running time is superior in this case.
 
\subsection{Parameter tuning on PMcut}
\begin{figure}[t!]
    \centering
    \begin{subfigure}{0.9\columnwidth}
                    \includegraphics[width=.99\columnwidth]{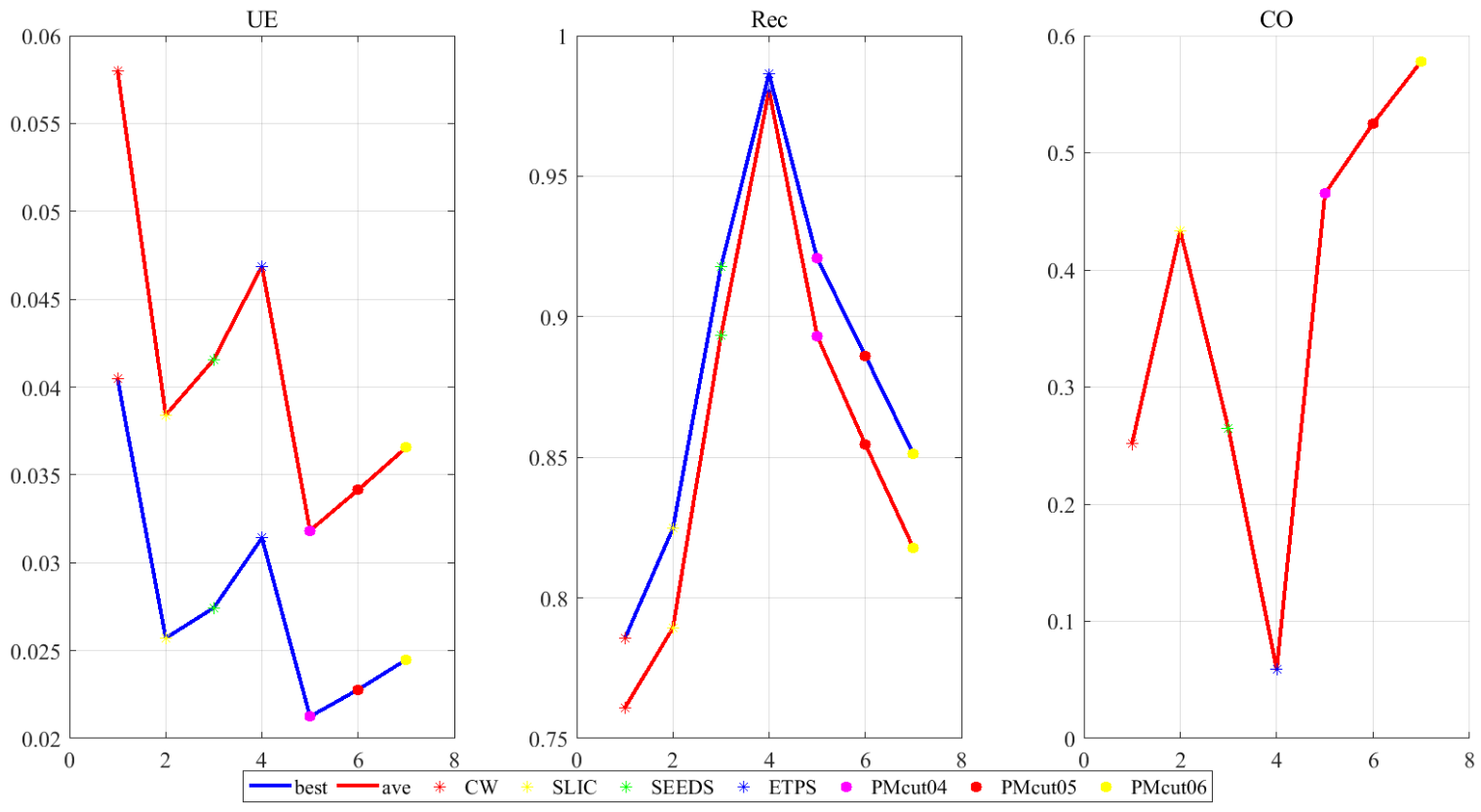}
    \end{subfigure}\\
    \begin{subfigure}{0.9\columnwidth}
            \includegraphics[width=.99\columnwidth]{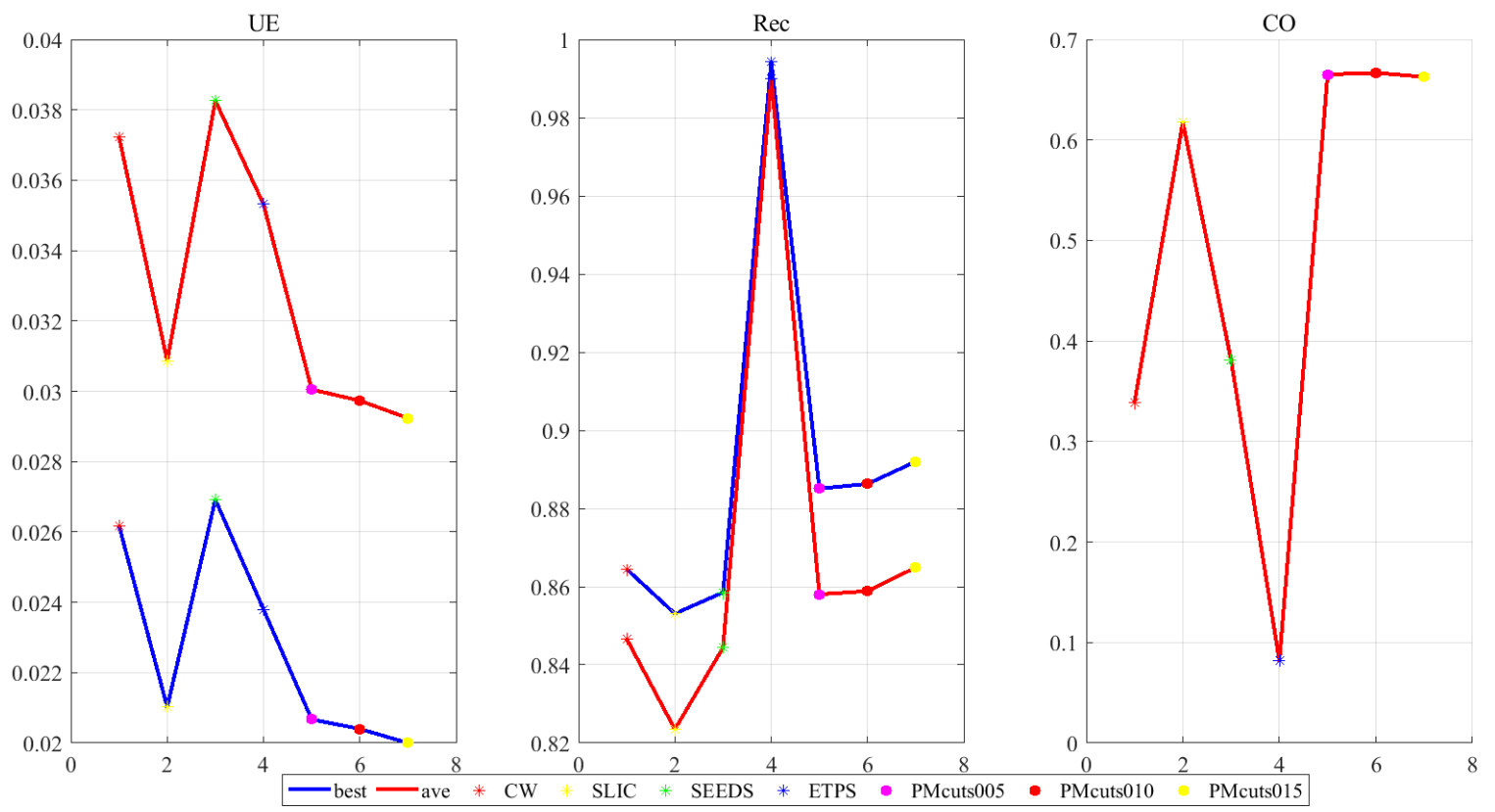}     
    \end{subfigure}       
     \caption{Top: PMcut on $S\&P$ images with different $\sigma$, $\#=1200$. Bottom: PMcut on Gaussian images with different time limit, $\#=1800$.}
     \label{pascal_test}
\end{figure}

Finally, we conduct experiments on adopting different values of regularization parameter $\sigma$ ($0.4$, $0.5$ and $0.6$) and time limit ($0.05$, $0.1$ and $0.15$ secs) within each patch of the image.
Fig.~\ref{pascal_test} shows two experiments on images with two kinds of noises.
It is obvious that when the regularization parameter $\sigma$ grows, Rec and UE become worse while CO better. Also, when the time limit increases, all three scores benefits, thus a trade-off between time and scores. We also report that the average MIP gap of adopting the $3$ time limits are $12.8\%$, $10.3\%$ and $9.5\%$.

\section{Conclusions}
\label{sec:conclusion}
We present a combined segmentation and denoising framework for superpixel generation specially on noisy images, where each rectangular patch of the original image consists of a MILP. We conduct extensive experiments on BSDS500 iamge dataset with noises. Our method achieves the best OP score, and exceeds the state-of-the-art on the UE and CO score.
Since image noise is unavoidable and there has been so far little work that addresses superpixels generation on noisy images, we believe this is a research direction that deserves the community's attention.

\bibliographystyle{splncs}
\bibliography{research}

\section{Appendix}
\subsection{Proof for Theorem $1$}

\begin{proof}
For any $x_e\in \x$, let $	\vartheta=\sum_{i\in V}|w_i-y_i |+\lambda \sum_{e\in E} x_{e}$ be the objective value of problem~\eqref{eq:Pott_1D2}, we proof two sufficient and necessary conditions of~\eqref{logic}.

\begin{enumerate}
	\item \label{enumerate1} $\nabla^1 w_e= 0 \Rightarrow x_{e} = 0.$

If $\nabla^1 w_e= 0$, by constraint~(\ref{bigmcons}, \ref{bigmcons2}), $x_e$ can be either $0$ or $1$. But the optimal solution will have $x_e=0$ since problem~\eqref{eq:Pott_1D2} is a minimization problem and $\lambda>0$, and this makes $\vartheta$ smaller.

	\item	$ x_{e} = 0 \Rightarrow \nabla^1 w_e= 0 .$
	
	If $x_e=0$, then it immediately follows by constraint~(\ref{bigmcons}, \ref{bigmcons2}) that $\nabla^1 w_e\leq 0$, and hence $\nabla^1 w_e= 0$.
	
	\item $\nabla^1 w_e\ne 0  \Rightarrow x_{e} = 1$.
	
	If $\nabla^1 w_e\ne 0$, it immediately follows by constraint~(\ref{bigmcons}, \ref{bigmcons2}) that $x_{e} =1$, and the assumpition that $M$ is big enough for~(\ref{bigmcons}, \ref{bigmcons2}) to hold.
	
	\item $x_{e} = 1  \Rightarrow \nabla^1 w_e\ne 0$.
	
	If $x_{e} = 1$, suppose we have $\nabla^1 w_e= 0$, then by proposition~\ref{enumerate1} of this proof, $x_{e} = 0$, thus a contradiction.
	
\end{enumerate}
\end{proof}

\subsection{Proof for Theorem $2$}
\begin{proof}
We prove this by stating that every $\x \in S_I^o$ satisfies the multicut constraints~\eqref{multicut_a}. 

As discussed in~\cite{kappesg}, the multicut constraints~\eqref{multicut_a} are equivalent to 
$$
\sum_{e\in C} x_e \neq 1, \;\forall \;\text{cycles}\; C\subseteq E.\nonumber
$$
Suppose there exists one solution $(\w,\x)\in S^o$ and a cycle $C^*=\{1,2,\ldots,n\}$ of length~$n-1$ (here, node $1$ and $n$ represent the same node) such that there is only one active edge $e = (i,i+1) \in C^*$ , $i\in [n-1]$ (i.e., $x_{i,i+1}=1$). By the definition of $S^o$, we have $w_i\neq w_{i+1}$.

Since all the other edges in cycle $C$ are dormant ($x_e=0$), by constraints~(\ref{bigmcons}, \ref{bigmcons2}), we have $w_1=w_2=\ldots=w_{i}$ and $w_i=w_{i+1}=\ldots=w_{n}$. Since node~$1$ and $n$ represent the same node, we have $w_1=w_2=\ldots=w_{n}$, thus $w_{i}=w_{i+1}$, and a contradiction. 
\end{proof}

\subsection{Proof for Theorem $3$}
\begin{proof}
Let $S_m$ be the set of all feasible solutions to the corresponding multicut problem~\eqref{multicut} with respect to the Potts~\eqref{eq:Pott_1D2}.
Since the chordless multicut constraints~\eqref{multicut_a} are facet-defining for the multicut polytope (convex hull of $S_m$), we just need to prove $S_I^o = S_m$.

From Theorem $2$, we know $S_I^o \subseteq S_m$. 
We prove $S_m  \subseteq S_I^o$ as follows. For any feasible solution $\x \in S_m$ of~\eqref{multicut}, it naturally defines a valid segmentation in the Potts model~\eqref{eq:Pott_1D2}. We can then compute the corresponding $\w$ by by fitting a linear function within each segment. Thus, we have constructed a feasible solution $(\w,\x)\in S^o$, which in turn implies the existence of $\x \in S_I^o$.
\end{proof}

\end{document}